\documentclass[]{Liebert_Author}

\title{ZodiAq: An Isotropic Flagella-Inspired Soft Underwater Drone for Safe Marine Exploration} 

\author
{Anup Teejo Mathew$^{1\ast}$, Daniel Feliu-Talegon$^{1}$, Yusuf Abdullahi Adamu$^{1}$, \\
Ikhlas Ben Hmida$^{1}$, Costanza Armanini$^{2}$, Cesare Stefanini$^{3}$,\\ Lakmal Seneviratne$^{4}$, Federico Renda$^{1,4}$\\
\\
 $^{1}$Department of Mechanical Engineering, Khalifa University, Abu Dhabi, UAE.\\
 $^{2}$Center for Artificial Intelligence and Robotics (CAIR), \\ New York University, Abu Dhabi, UAE\\
 $^{3}$Department of Excellence in Robotics and AI, The Biorobotics Institute,\\  Scuola Superiore Sant’Anna, Pisa 56127, Italy\\
 $^{4}$Khalifa University Center for Autonomous Robotic Systems (KUCARS), \\
 Khalifa University, Abu Dhabi, UAE\\
$^\ast$Corresponding author: anup.mathew@ku.ac.ae
}
\usepackage{endfloat}
\usepackage{pdfpages}
\usepackage{placeins}
\usepackage{bm}
\usepackage{hyperref}
\hypersetup{
    colorlinks=true,
    linkcolor=blue,
    filecolor=magenta,      
    urlcolor=cyan,
    pdftitle={Overleaf Example},
    pdfpagemode=FullScreen,
    }
\usepackage[a4paper, margin=2cm]{geometry}

\begin{document} 
\includepdf[pages=-]{Notice.PDF}
\maketitle 


\begin{abstract}
  The inherent challenges of robotic underwater exploration, such as hydrodynamic effects, the complexity of dynamic coupling, and the necessity for sensitive interaction with marine life, call for the adoption of soft robotic approaches in marine exploration. To address this, we present a novel prototype, ZodiAq, a soft underwater drone inspired by prokaryotic bacterial flagella. ZodiAq's unique dodecahedral structure, equipped with 12 flagella-like arms, ensures design redundancy and compliance, ideal for navigating complex underwater terrains. The prototype features a central unit based on a Raspberry Pi, connected to a sensory system for inertial, depth, and vision detection, and an acoustic modem for communication. Combined with the implemented control law, it renders ZodiAq an intelligent system. This paper details the design and fabrication process of ZodiAq, highlighting design choices and prototype capabilities. Based on the strain-based modeling of Cosserat rods, we have developed a digital twin of the prototype within a simulation toolbox to ease analysis and control. To optimize its operation in dynamic aquatic conditions, a simplified model-based controller has been developed and implemented, facilitating intelligent and adaptive movement in the hydrodynamic environment. Extensive experimental demonstrations highlight the drone's potential, showcasing its design redundancy, embodied intelligence, crawling gait, and practical applications in diverse underwater settings. This research contributes significantly to the field of underwater soft robotics, offering a promising new avenue for safe, efficient, and environmentally conscious underwater exploration.
\end{abstract}

\section{Introduction}

Our vast ocean, covering more than 70\% of the Earth's surface, is a realm of unparalleled potential and critical environmental concern. These waters, teeming with diverse ecosystems, are vital not only for marine life but also for the health of the planet and the well-being of billions of people. 
In this context, the development and deployment of unmanned underwater Remotely Operated Vehicles (ROVs) are of immense significance. While there exist several industrial solutions for tasks such as environmental monitoring and resource mapping \cite{Petillot2019}, the closeup exploration of underwater life and navigation near cluttered seabeds or around submerged structures demand bio-inspired designs that can prevent unintended impacts with the delicate marine environment \cite{Katzschmann2018}. The high-frequency rotation of traditional motors close to the sea floor or marine life can cause self-damage and disturb marine life, making it harder to study underwater ecosystems.
\par

Compared to their rigid counterparts, the compliant nature of soft robots allows for safer, less precision-demanding, bio-compatible, and adaptable interactions with unpredictable environments. Inspired by the efficient morphology and body structure of aquatic organisms, soft roboticists have proposed a range of bio-inspired designs for different swimming strategies \cite{Aracri2021, Li2023b, Qu2023, Calisti2017RS}.
Lift-powered swimming, inspired by the techniques of animals such as manta rays \cite{Chen2012} and turtles \cite{Kim2013}, utilizes hydrodynamic lift forces generated by the movement of their fins or limbs. The undulation strategy, in contrast, relies on the body's wave-like motion, effectively producing propulsion thrust through the cyclical storage and release of elastic energy. Robot designs in this category draw inspiration from the slender and streamlined bodies of creatures like snakes \cite{Kelasidi2016}, and eels \cite{Nguyen2022}, making them ideal for navigating narrow environments. Animals such as fishes, frogs, and insects utilize limb movements reminiscent of oars, pushing water backward to generate forward thrust from the resulting reaction force. This drag-powered swimming gait is characterized by its agility and reduced turning radius and has been adapted in various robotic designs \cite{Katzschmann2018, Jia2015, Marchese2014}. Lastly, the jet propulsion strategy, used by animals like squids and octopi, propels them forward by expelling water, utilizing the conservation of momentum principle \cite{Godaba2016, Giorgio-Serchi2016}.
\par 
Most soft robot designs, inspired by the aforementioned locomotion strategies, require intricate mechanisms or the incorporation of smart materials. This contrasts with the simpler, conventional motor-based propellers used in rigid underwater vehicles.
Recently, a novel macroscale underwater locomotion strategy has been developed in a high Reynolds number regime, combining motor rotary actuation with an elongated passive soft filament acting as the rotor \cite{Calisti_IROS_2019}.
The strategy mimics the swimming gait of bacterial flagella, often regarded as the only example of a biological ``wheel."
The transmitted torque and external fluid interaction on the soft body prompt the soft body to adopt a helical shape, effectively generating the propulsion thrust needed for movement \cite{Yu2006, Armanini2022RAL}. Unlike traditional (rigid) propellers that depend on high-frequency, low-torque rotation, these macroscale soft propellers utilize low-frequency, high-torque actuation, offering a powerful yet biocompatible mode of locomotion.  
In a previous work \cite{Armanini_TRO2021_2}, we presented a comprehensive design, characterization, and modeling framework of flagellate underwater robots.
The flagella were modeled using the strain-based modeling approach of Cosserat rods \cite{Renda_RAL2020, Boyer_TRO2020}, while simplifying assumptions are used to model fluid interaction forces due to buoyancy, drag, lift, and added mass.
%
%
%
%

\par 
Underwater drones used for close inspection of marine environments can benefit from a compact turning radius and the ability to maneuver in all three dimensions. Designing with redundancy is beneficial, allowing the drone to remain functional even if some components fail, greatly improving its reliability in complex underwater settings.
Accounting for all these functionalities, we introduce a new design for a soft robot comprising 12 flagella modules arranged symmetrically across a dodecahedral shell. We named it ``ZodiAq" to signify its characteristic shape with 12 actuators, with `Aq' indicating its aquatic nature.
%
The design and fabrication details of the prototype are provided in Section \ref{sec::Design&Fabrication}.
\par
To analyze locomotion and develop control laws, we created a digital twin of ZodiAq using SoRoSim, a MATLAB toolbox based on the Geometric Variable Strain (GVS) model \cite{Mathew2022}.
The methodology for constructing the digital twin and a sample simulation result are described in Section \ref{sec::Model}. We have developed a simplified model-based control law and implemented it on the digital twin to control all 6 DoFs of the robot body, while for the prototype, control of depth and orientation is achieved.
Details of the control law, simulation, and experimental results are described in Section \ref{sec::control}.

\par 
Using four experimental demonstrations (Section \ref{sec::demo}), we showcase the capabilities of ZodiAq and their real-world applications. Our first demonstration focuses on the design redundancy of ZodiAq, which ensures that the prototype remains functional even with the failure of an actuator module.
In the second demonstration, we explored ZodiAq's 'Embodied Intelligence,' derived from its soft body's physical form (compliance) \cite{Pfeifer2007, Mengaldo2022}. This built-in intelligence facilitates seamless interaction with the environment, minimizing control needs and optimizing the use of environmental features.
The third demonstration is dedicated to showing another locomotion gait of the prototype: crawling. Finally, the fourth demonstration highlights ZodiAq's ability to navigate artificial coral, showcasing its suitability for close inspection tasks in marine settings.
%
%
The final section (Section \ref{sec::conclusion}) provides conclusive remarks on the results presented in this work and outlines potential future developments of the proposed system.

\section{Design and Fabrication}
\label{sec::Design&Fabrication}
\begin{figure}
\centering
\includegraphics[width=1\columnwidth]{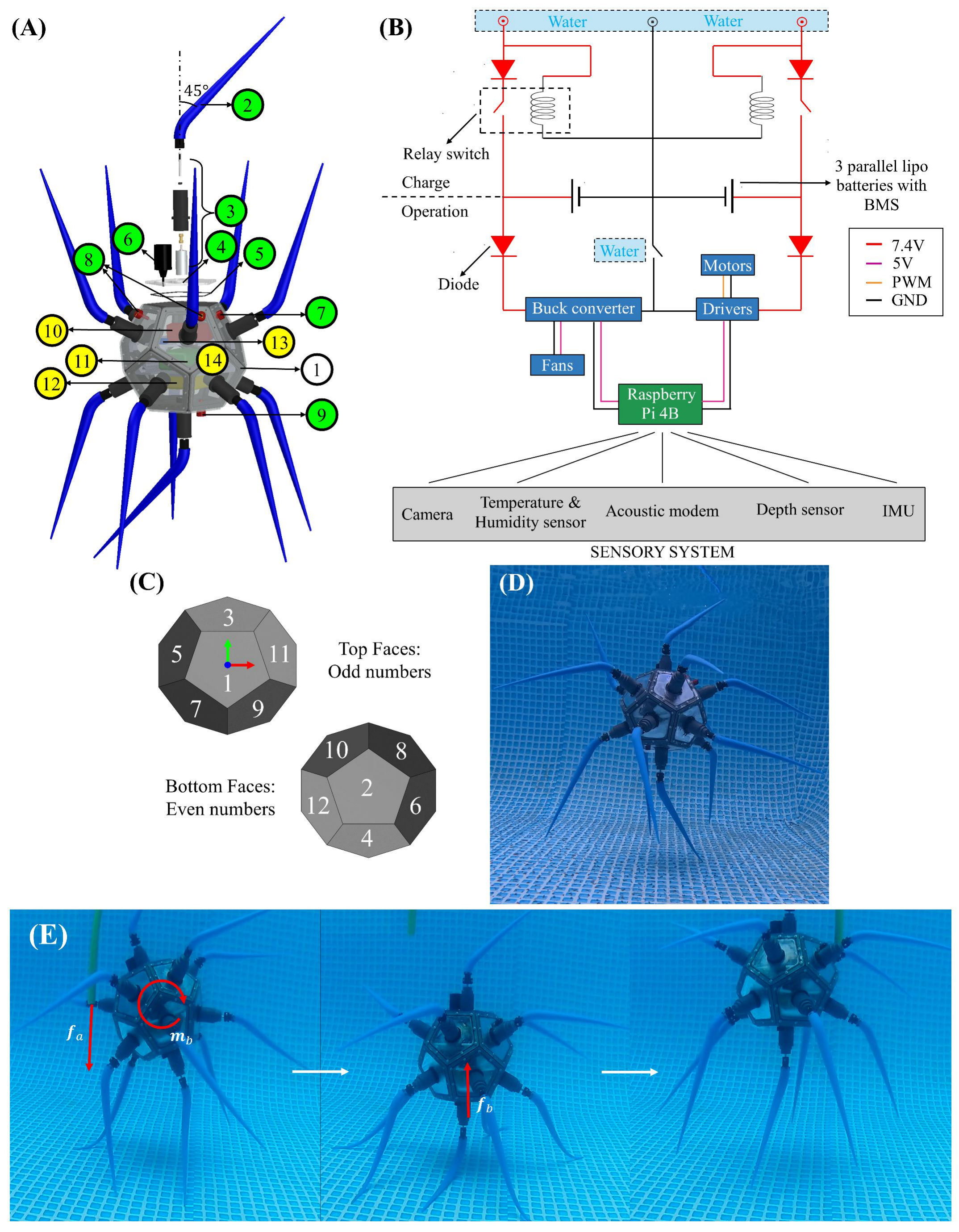}
\caption{Prototype Design: (A) CAD model of the prototype (including an exploded view of one face of the prototype), with external components marked in green and internal components in yellow. Components are listed in Table \ref{tab::ZodiAq Components}, (B) Electrical connection between components, (C) Numbering scheme used for the faces and motors, (C) Fully assembled prototype floating in the water, (E) a Demonstration of passive mechanical stabilization due to neutral and stable buoyancy. A force ($f_a$) is applied to tilt and push the Zodiaq down. The resulting moment ($m_b$) due to the arrangement of CB above the CG corrects the tilt as the robot sinks deeper. The force of buoyancy ($f_b$) restores the height.
}
\label{fig::Design}
\end{figure}

To achieve omnidirectional movement and rotation about the vertical axis, as well as to ensure design redundancy, we opted to equip the robot with 12 flagella modules. For geometrical symmetry, we chose a dodecahedron – the fourth Platonic solid – as the shell structure of the robot (Figure \ref{fig::Design}A). Each pentagonal face of the robot houses a motor canister assembly to power the flagella, which are fabricated through a silicone molding process with a 3D-printed hook embedded at their base.
ZodiAq utilizes a Raspberry Pi to control the angular speed and direction of each DC motor. The sensory system of ZodiAq consists of a camera for vision, sensors including a temperature and humidity sensor for internal damage detection, an Inertial Measurement Unit (IMU) for motion detection, a depth sensor for depth feedback, and an acoustic modem for underwater communication. The entire system is powered by Lithium Polymer (LiPo) batteries, which are mounted inside. Lastly, a ballast is mounted inside to shift the center of gravity below the center of buoyancy. A comprehensive list of all external and internal components of ZodiAq are listed in Table \ref{tab::ZodiAq Components}.

\begin{table}[h!]
\begin{threeparttable}
\caption{Components of ZodiAq according to Figure \ref{fig::Design}A
\label{tab::ZodiAq Components}}
\begin{tabular}{@{}ll}
\toprule
 Number  & Description  \\
\midrule
  1  & Dodecahedral frame (PVC) and rack (PLA)   \\
  2  & Flagella with Hook $\times 12$ (Silicone: Mold Star™ 30)   \\
  3  & Motor canister assembly$^{a}$ $\times 12$   \\
  4  & Faceplate (Polycarbonate) $\times 12$ \\
  5  & Face O-rings $\times 12$  \\
  6  & Acoustic Modem (Succorfish Nanomodem v3)   \\
  7  & ON/OFF switch   \\
  8  & Charging ports $\times 2$  \\
  9  & Depth Sensor   \\
  10  & Raspberry Pi 4B and accessories$^{b}$   \\
  11  & Set of three 7.4V 2S Lipo batteries $\times 2$   \\
  12  & Vinyl coated Lead ballasts (4 kg)   \\
  13  & Small DC fan $\times 2$   \\
  14  & Other electrical components$^{c}$   \\
\bottomrule
\end{tabular}
\begin{tablenotes}[flushleft]\footnotesize
\item[${a}$] Subassembly of DC motor, shaft coupler, shaft, shaft O-rings, and a PVC canister.
\item[${b}$] Adafruit DC motor HATs (Motor drivers) $\times 3$, IMU, Temperature \& Humidity sensor, and PiCamera
\item[${c}$] Buck converter (step-down transformer), relay switches, battery management system (BMS), voltage indicators, electrical connections
\end{tablenotes}
\end{threeparttable}
\end{table}

The schematic of the electrical connection is shown in Figure \ref{fig::Design}B. The charging and operation circuits are split into two: one for the Raspberry Pi and another for the motor drivers. The setup includes components for short-circuit protection, voltage balancing, and voltage conversion. The motors associated with the 12 flagella modules follow a numbering scheme mentioned in Figure \ref{fig::Design}C. Pairs of consecutive motors (1 and 2, 3 and 4, etc.) are arranged on opposite sides of the prototype. This particular arrangement ensures minimum load on the motor drivers and assists in simplifying the control law (details in the Supplementary Material). 

The components are methodically arranged on two internal racks: the lower rack houses batteries and ballasts, while the Raspberry Pi and its peripherals are mounted on the upper rack. The mass of all the internal components is balanced against the water mass of the displaced volume of the ZodiAq's shell. The mass distribution and balancing results in the prototype's neutral and stable buoyancy. The fully assembled ZodiAq prototype, immersed underwater, is shown in Figure \ref{fig::Design}D.
\par
We tested the prototype's functionality up to a depth of 2.5 meters and an estimated operational duration of 1 hour, constrained primarily by Raspberry Pi's energy consumption. The neutral stability of the prototype is validated in an experiment illustrated in Figure \ref{fig::Design}D. In this test, a vertical force ($f_a$) was applied to a side-mounted motor canister, causing the prototype to descend and tilt. The force ($f_b$) and the stabilizing moment ($m_b$) due to buoyancy successfully repositioned the prototype, restoring its height and orientation. Readers may refer to the video demonstration in the supplementary movie S1. Table \ref{tab::ZodiAq Specifications} lists the key specifications of the prototype. For detailed information on the design and fabrication, readers may refer to the supplementary material.

\begin{table}[h!]
\begin{threeparttable}
\caption{Key specifications of ZodiAq
\label{tab::ZodiAq Specifications}}
\begin{tabular}{@{}ll}
\toprule
 Property & Value  \\
\midrule
  Edge length of the dodecahedral shell & 10 cm  \\
  Shell mass & 8.57 kg  \\
  Total mass & 10.75 kg \\
  Downward shift in CG ($d_{CG}$) & 3.5 cm   \\
  Body Length ($BL$) (excluding flagella) & 30 cm   \\
  Maximum operating time  & 1 hour  \\
  Maximum motor speed ($\omega_{max}$) & 130 RPM  \\
  Maximum translational (x-, y-, and z-axis) speed & 8 $BL$/min\\
  Maximum rotational (about z-axis) speed & $3\pi$ rad/min\\

\bottomrule
\end{tabular}
\end{threeparttable}
\end{table}

\section{Digital Twin of ZodiAq}\label{sec::Model}
To analyze the dynamics and formulate control laws, we created a digital twin of ZodiAq using the SoRoSim toolbox. The toolbox was developed based on the Geometric Variable Strain (GVS) approach, offering a unified mathematical framework for modeling soft and rigid bodies \cite{Mathew2022}. The GVS approach is based on the strain-parameterization of the Cosserat rod: a 1D continuum mechanics object with axial torsion, bending in two directions, axial stretch, and shear in two directions. A summary of the GVS model is provided in the supplementary material, while for a detailed review, the readers may refer to \cite{Renda_RAL2020, Boyer_TRO2020}. The versatility of SoRoSim allows for the analysis of open-, closed-, and branched hybrid robots in various external load and actuation conditions, making it ideal for modeling ZodiAq. 

\begin{figure}
\centering
\includegraphics[width=1\columnwidth]{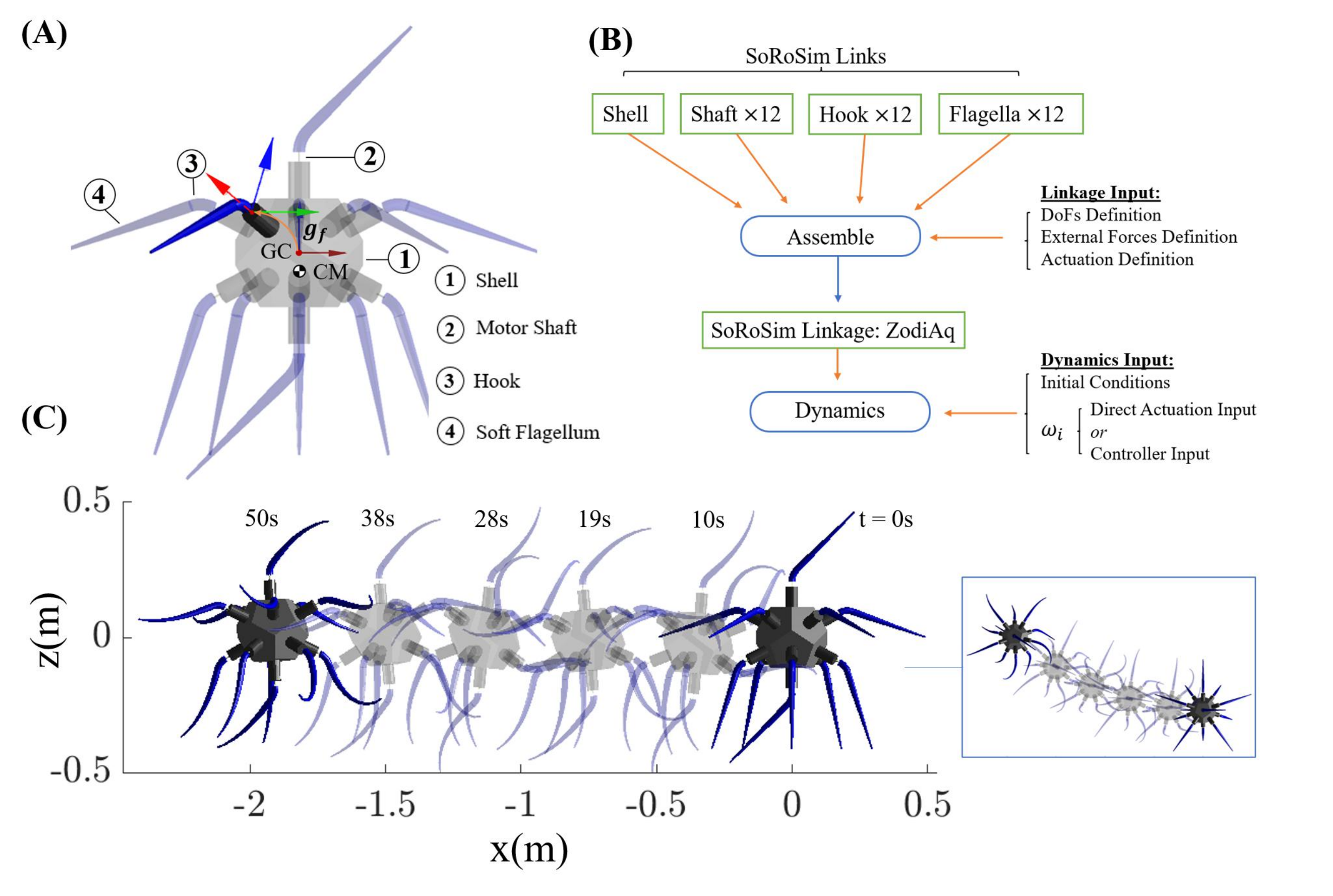}
\caption{(A) Digital twin of the robot, created using SoRoSim toolbox. GC is the Geometric Centre, CM is the Centre of Mass, and $\bm{g}_f$ is the transformation matrix from the GC to a motor shaft (M7 in the diagram). (B) A simplified flowchart showing the processes in SoRoSim. (C) Superimposed images of the robot at different time steps during a dynamic simulation. An inset shows a top view (xy-plane) of the motion}
\label{fig::Model}
\end{figure}

The toolbox defines links and their assembly, linkages, as MATLAB class objects. Each link has a rigid joint and a soft or rigid body. The geometric and material properties of the link are defined during link creation. For the robot's dodecahedral shell, we created a rigid link with a 6 DoF free joint. Shafts are created as rigid links with revolute joints and assembled appropriately using a fixed transformation matrix $\bm{g}_f$ (Figure \ref{fig::Model}A) corresponding to each face of the shell. Connected to each shaft is a hook, a rigid link with a fixed joint (0 DoF) and a pre-curvature, and to each hook we connect a flagellum, a soft link with a fixed joint.
We model the soft flagellum as an inextensible Kirchhoff rod, a sub-class of Cosserat rod with torsion and bending with linear strain parameterization (6 DoFs).
The total number of links in the linkage (assembly) is 37 and the total DoFs of the system are 90.
The linkage creation also allows users to define external forces and actuators. External forces due to gravity and fluid interaction including buoyancy, drag, lift, and added mass (fluid displacement) are considered. Joints of the shafts are defined as actuators that are controlled by joint angles. Finally, a simplified control law, discussed in the next section, is implemented. The overview of the linkage creation in SoRoSim is shown in Figure \ref{fig::Model}B. For a comprehensive understanding, readers may refer to the supplementary material.

Analysis of dynamic response is a `method' of the linkage class object. For an illustrative simulation, we arbitrarily actuated motors M6, M8, M9, and M11, which are distributed predominantly on one side of the dodecahedral shell, at 60 RPM in the CCW direction (Figure \ref{fig::Model}C). We noted that besides the intended straight-line motion towards the opposite side (-ve x-axis), there was an unintended drift of $0.74 m$ in the y-axis and a minor displacement of $0.03 m$ meters in the z-axis. Additionally, the robot experienced a significant orientation shift about the z-axis by $32.8^{\circ}$. The deployment of controllers, which will be discussed in the subsequent section, could aid in sustaining precise orientation and trajectory during movement. The dynamic simulation video is available for viewing in the supplementary movie S2. Readers can also simulate the digital twin in MATLAB through the GitHub link provided in \cite{ZodiAqGitHub}.


\section{Nonlinear Motion Control for Robot Navigation}

\label{sec::control}
The dynamic modeling of our robot is highly complex, primarily due to the interaction between the soft parts with the fluid. With a total of 90 DoFs and 12 inputs, the dynamics of the robot presents substantial challenges for model-based control design. We propose a simplified model that solely accounts for the robot's shell. The collective input, or throttle input, is determined by aggregating the forces and moments generated by each flagellum positioned at the center of each face. The proposed model allows, quite realistically, to develop a nonlinear model-based controller of the system for performing robot navigation. The control law demonstrates the tracking performance in the simplified system's states, encompassing the 3D position and orientation about the vertical axis ($\psi$) of the shell. The orientation of the remaining two angles stabilizes towards the zero equilibrium point, showcasing the system's inherent stability and highlighting the control over the six DoFs of the main body in simulations. The developed control law has been employed in the complete system dynamics (digital twin), demonstrating the effectiveness of the simplified model for control design.  However, due to the absence of a tracking system in our experimental setup, planar coordinates ($x$ and $y$) feedback is unavailable for the actual prototype. Consequently, our control implemented in the real prototype is limited to depth ($z$) and orientation about the vertical axis ($\psi$) while the planar movement (x and y) was controlled in
an open-loop manner by commanding a desired acceleration. This adjustment in the control law enables the execution of various trajectories with the actual prototype, including straight-line movements, square paths, and rotational maneuvers, as we demonstrate later on. In the experiments, the depth measurement is directly obtained from the depth sensor, while the orientation angle is derived from the calibrated magnetic field measurements provided by the IMU. The details of the simplified model and the control law are explained in Supplementary Material.


\subsection{Closed-Loop Control Simulation}

\begin{figure}
\centering
\includegraphics[width=1\columnwidth]{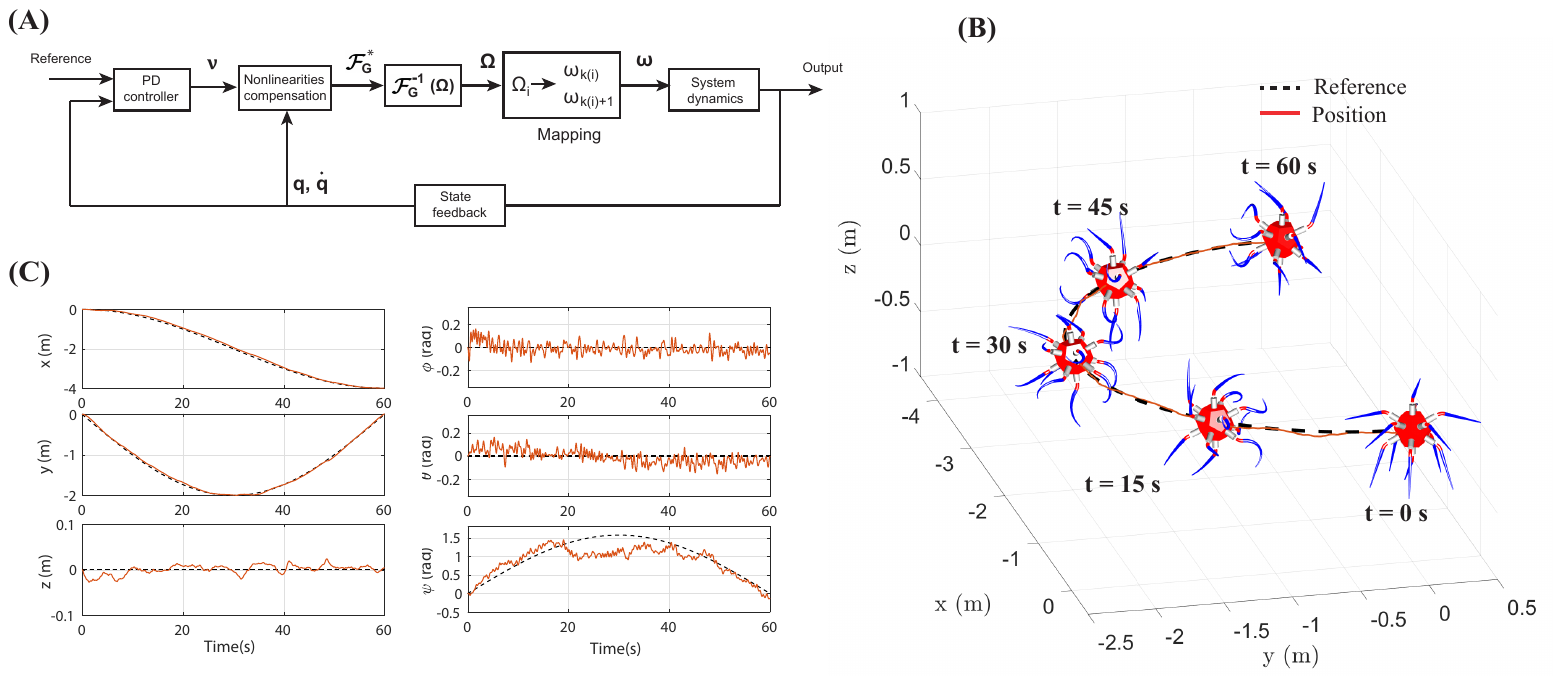}
\caption{Simulated results of the closed-loop system : (A)  Schematic illustrating the control strategy implemented in the closed-loop system. The control law aims to simultaneously control the 3D position ($x,y,z$) and orientation ($\psi$). The angular speed of the motors, denoted as $\bm{\omega}$, serves as the system's control input. The output $\bm{q}=[\phi,\theta, \psi, x, y, z]^{T}$ represents the generalized coordinates of the system, while $\bm{\nu}$ represents the output of proportional-derivative (PD) controller. $\bm{\mathcal{F}}_{G}^{*}$ denotes the desired resultant forces and moments applied to the CM, and $\bm{\Omega}$ is the fictitious input that renders the simplified model differentially flat. For a detailed explanation of the simplified model and the proposed control law, readers can refer to the supplementary material. (B) Comparison between reference trajectories and the actual positions of the angular and linear coordinates. (C) 3D representation of the simulation.}
\label{fig::Control}
\end{figure}

The effectiveness of the proposed control method is tested using the digital twin, which simulates the whole dynamics of the system. The proposed control law predefined the rotation direction of each flagellum, considering that the direction of the rotation does not impact the direction of the thrust. Moreover, it combines the two motors of opposite faces to create a new input, making the system differentially flat. We conduct closed-loop control simulations where the system executes a planar semicircular trajectory with a radius of $2$ m over $60$ s while also controlling the orientation of the vertical axis $\psi$ and the depth $z$ of the system. Moreover, we can see that the Euler angles $\phi$ and $\theta$ remain stable around the equilibrium point without the need to control those variables directly in the closed loop.  The simulation results, together with the developed control scheme, are shown in Figure \ref{fig::Control}.  Readers may watch the supplementary movie S3 for a dynamic simulation of ZodiAq demonstrating semicircular trajectory tracking.

\subsection{Depth and Orientation Control}

\begin{figure}
\centering
\includegraphics[width=1\columnwidth]{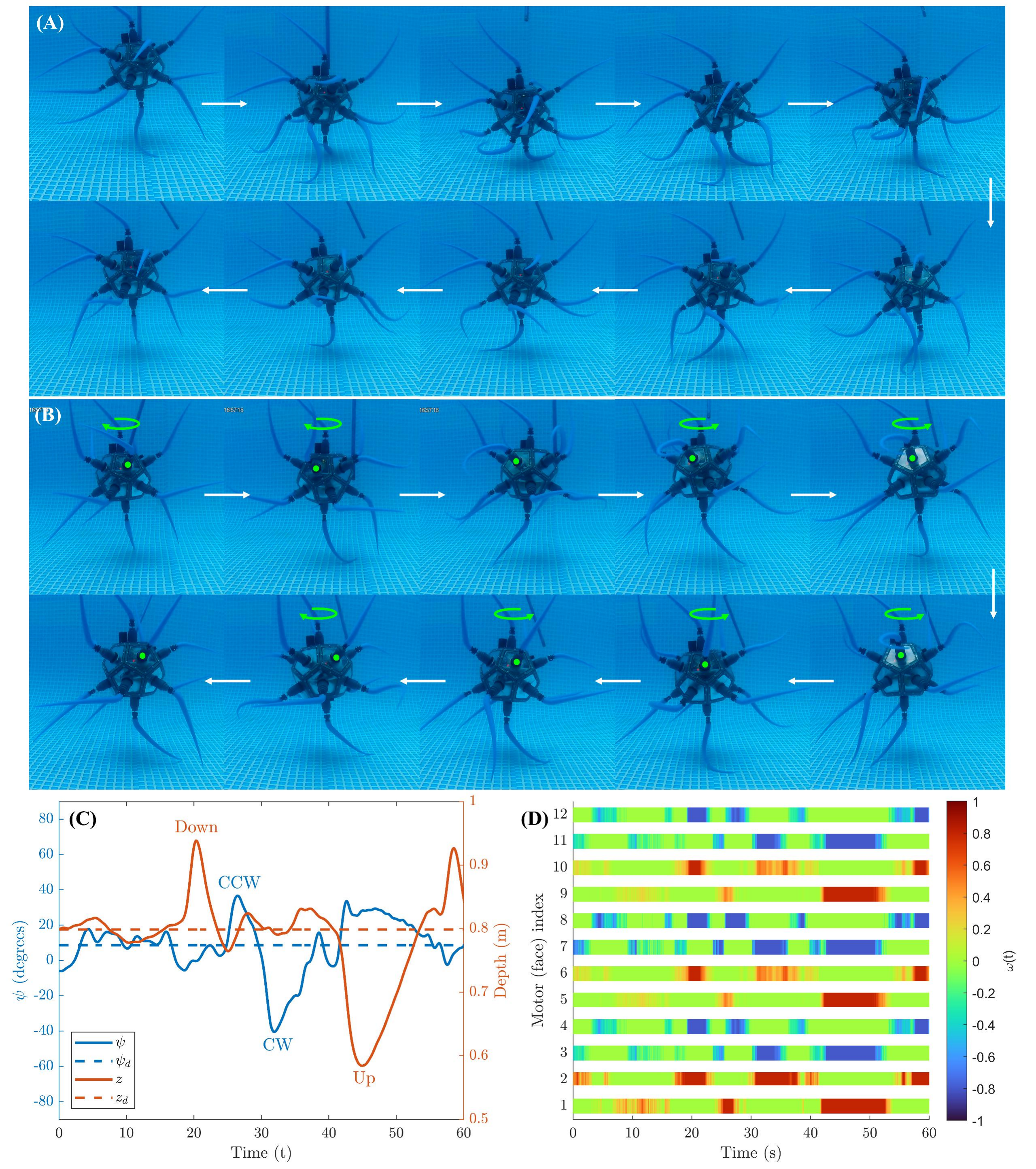}
\caption{Validation of Height and Orientation Control: (A) The sequence of video snapshots illustrates the active depth control by controlling the motors using depth feedback. (B) Process of orientation stabilization using IMU feedback. The direction of the ZodiAq’s rotation is shown by green arrows, while the green dot on a face assists in visualizing the orientation throughout the maneuver. (C) Measured angle and depth and their desired values showing controller in action during various experiment instances: pushing down, counterclockwise (CCW) rotation, clockwise (CW) rotation, and pulling up. (D) Controlled motor speed ($\omega/\omega_{max}$) vs time. The color of the band represents the value of angular velocity.
}
\label{fig::HeightOrientation}
\end{figure}

Our preliminary experiment focused on validating the control of height and orientation (Figure \ref{fig::HeightOrientation}). The desired orientation and depth were established based on the prototype's initial measurements upon being submerged. Once the robot achieved stability, the prototype's depth and orientation were intentionally altered by applying external forces and moments. The control system responded adeptly to these disturbances, effectively restoring the prototype to its pre-set depth and orientation. Figure \ref{fig::HeightOrientation}A demonstrates the depth control: the drone is pushed downward using a rod. Subsequently, using the depth feedback, the control system actively minimizes the depth error and returns the drone to its initial depth. Notably, the buoyancy force also plays a role in facilitating this correction. Figure \ref{fig::HeightOrientation}B illustrates the orientation control: after a forced clockwise rotation of the drone, the control system corrects the induced orientation error, realigning to the initial orientation. 
Readers can access the supplementary movie S4 to see a video demonstration of height and orientation control, including CW and CCW rotations, as well as vertical shifts downwards and upwards. Figure \ref{fig::HeightOrientation}C shows the measured and desired values of orientation angle and depth, while Figure \ref{fig::HeightOrientation}D illustrates the output rotational velocities ($\omega/\omega_{max}$) of all twelve motors, as determined by the controller. To avoid overloading the motor driver, the velocities have been capped at a magnitude of $0.8\omega_{max}$. The restoring action of the controller can be appreciated from the velocities applied to each motor at different instances of the experiment.

\subsection{Square Trajectory}
The second experiment aimed to navigate a square trajectory. Due to the absence of planar coordinate feedback in our experimental setup, this method involved maintaining initial orientation and depth while movement along the x- and y-axes are regulated by commanding a desired acceleration. The procedure entailed commanding a positive acceleration in the x-direction and then a positive acceleration in the y-direction. This was followed by commanding a negative acceleration in the x-direction and, finally, a negative acceleration in the y-direction, with each step lasting for 40 seconds.

\par
\begin{figure}
\centering
\includegraphics[width=1\columnwidth]{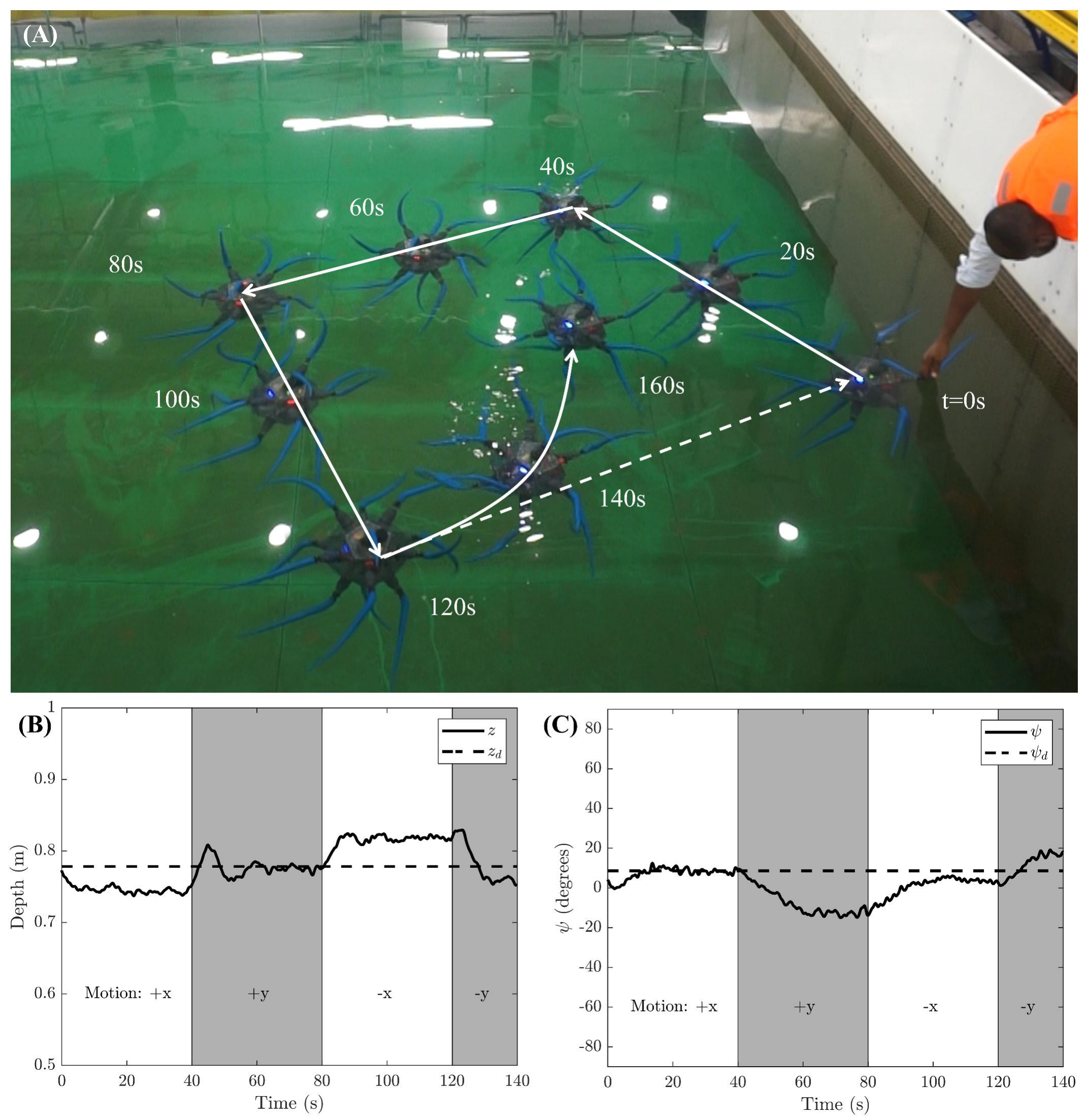}
\caption{Square Trajectory in Open-loop: (A) Superimposed images of the drone attempting to follow a square trajectory in open-loop control. (B) Robot depth measurement obtained from the depth sensor. (C) Orientation about the vertical axis.
}
\label{fig::SquareTrajectory}
\end{figure}

Figure \ref{fig::SquareTrajectory}A shows the experimental result of following a square trajectory. Readers may refer to the supplementary movie S5 for the video result. Directions of motion are determined by the initial orientation of the prototype. The prototype successfully follows the square trajectory, except on the fourth side, where a notable deviation is observed. This deviation can be attributed to imperfections in the prototype and the absence of planar coordinates feedback. Additionally, it is important to note that there is no feedback mechanism for planar coordinates. The maximum error in the height is 5 cm (Figure \ref{fig::SquareTrajectory}B), while the maximum orientation error is 23 degrees (Figure \ref{fig::SquareTrajectory}C).

\section{Experimental Demonstrations}
To demonstrate the prototype's functionalities and potential uses, we delve into experimental demonstrations of design redundancy, embodied intelligence, crawling locomotion, and its application in close exploration of marine environments. Readers may refer to supplementary videos S6 through S9 for video results of these demonstrations.
\label{sec::demo}
\subsection{Prototype Redundancy}
\begin{figure}
\centering
\includegraphics[width=1\columnwidth]{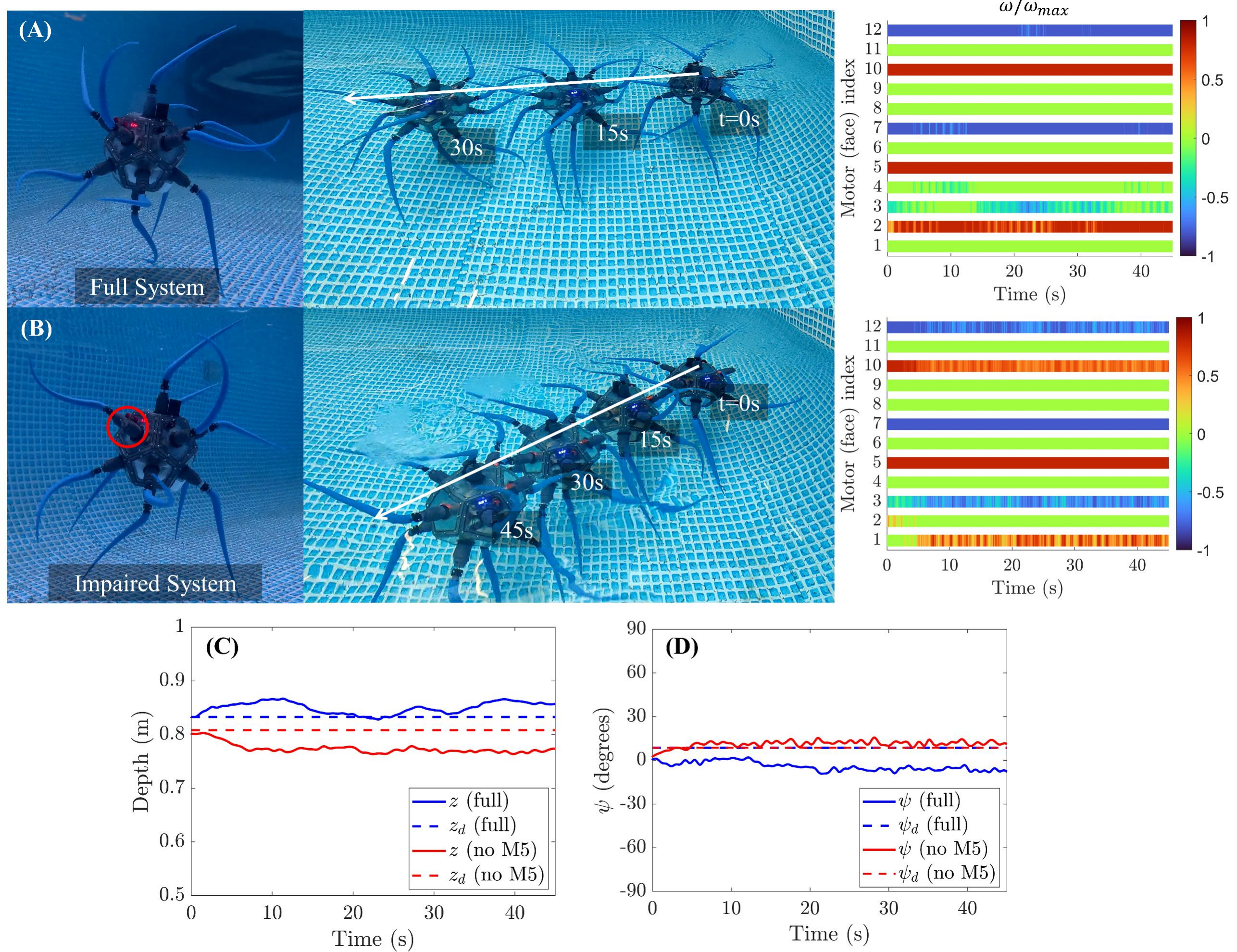}
\caption{Assessing Design Redundancy: (A) Full system executing a straight-line trajectory. Note that the motors M2, M5, M7, M10, and M12 were most active. (B) The impaired system (after removing flagella attached to M5) attempts the straight-line trajectory. M1 is activated, and M3's activity is increased to compensate for M5's missing flagella. The controller still engaged M5, unaware of the impairment. (C) Comparison of depth measurements for both full and impaired systems. (D) Comparison of orientation measurements for both scenarios
}
\label{fig::Redundancy}
\end{figure}

Many factors can lead to the damage of a robotic system as it navigates a complex and unknown terrain. The ability of the robot to accomplish given tasks having lost the functionality of some part, a severed tentacle or defective motor, for instance, defines the robot’s redundancy. This important feature enhances the adaptability and reliability of the robots that operate in unpredictable underwater environments \cite{Liu2023}. The ZodiAq is equipped with 12 actuators specifically to ensure operational redundancy. When an actuator becomes damaged, the robot should redistribute the affected operation among other actuators, significantly lowering the risk of critical failures. To test the redundancy of ZodiAq, we conducted the following experiments (Figure \ref{fig::Redundancy}).
\par
In the first experiment, ZodiAq was set to move in a straight line in the positive x-direction (acceleration command) with all flagella intact (full system). In the second experiment, we removed one of the actuators (M5), thereby creating an impaired system.
As illustrated in Figure \ref{fig::Redundancy}A, the full system covered a distance of two body lengths every 15 seconds. The impaired system also maintained a straight-line trajectory as depicted in Figure \ref{fig::Redundancy}B. However, its speed was reduced to one body length every 15 seconds, nearly half the velocity achieved by the full system. It's important to note that the initial orientation and depth were established at the time of ZodiAq's placement, resulting in a different x-axis direction for each experiment.
From the plot of controlled motor speed against time, the variations in motor speed for both cases can be inferred.
%
Figures \ref{fig::Redundancy}C and D provide insights into the depth (z) and orientation ($\psi$) of the robot throughout the experiment. Despite the impairment, the system successfully maintained the measured depth and orientation close to the desired values. 

\subsection{Embodied Intelligence}
\begin{figure}
\centering
\includegraphics[width=1\columnwidth]{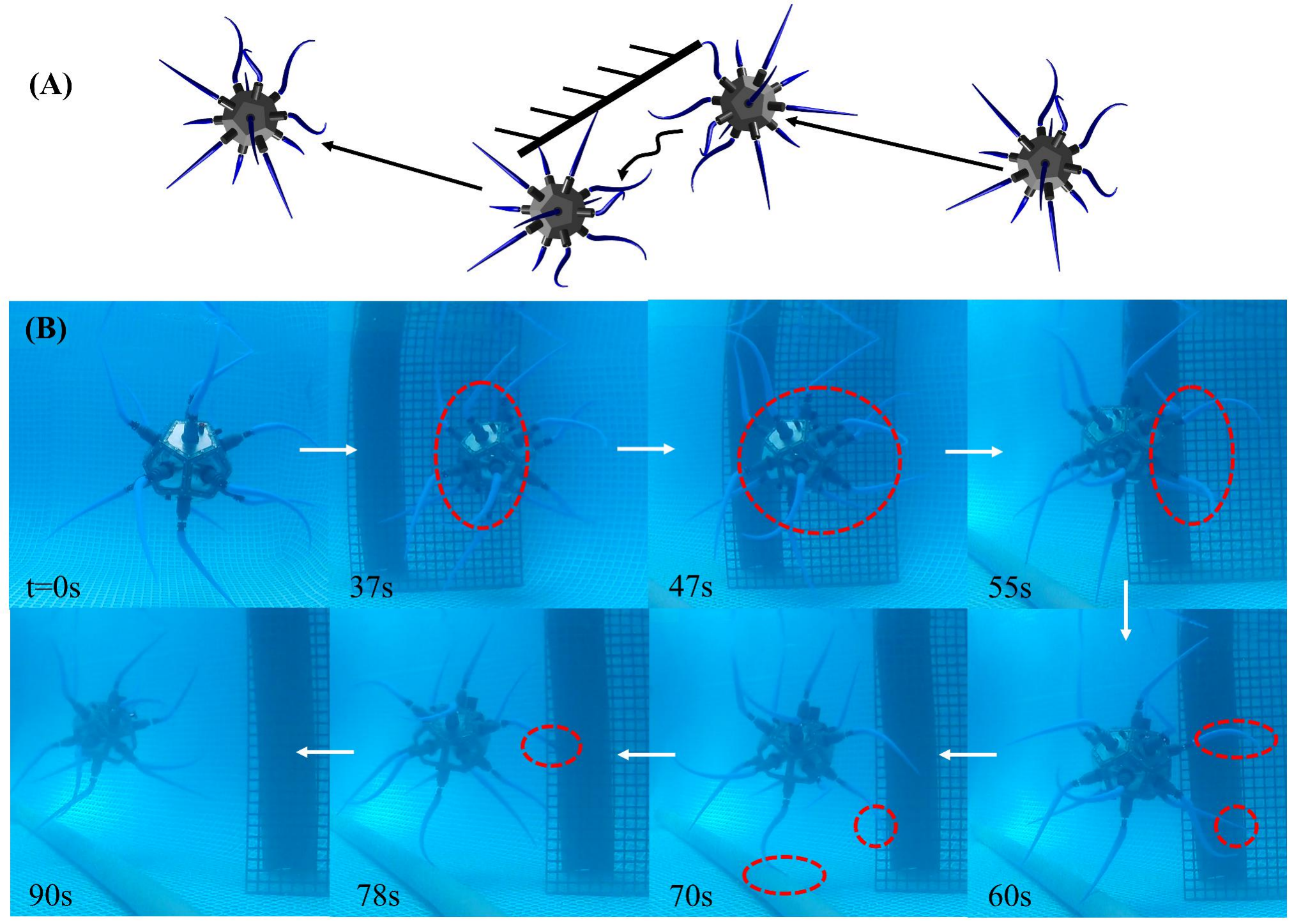}
\caption{Demonstration of Soft Interaction and Adaptive Trajectory of the ZodiAq as it follows an open loop straight line motion: (A) Schematics of the scenario. (B) Snapshots of the experimental demonstration capture the robot's soft flagella engaging with a slanted barrier. Contact areas are demarcated with dashed ellipses.
}
\label{fig::SoftInteraction}
\end{figure}

In recent times, compliant structures that are capable of using the environment to their advantage rather than avoiding them are getting attention from the soft robotics community \cite{Pfeifer2007, Mengaldo2022}. This is in response to the need for robot platforms that can safely interact in complex and dynamic surroundings, especially through embodied cognition \cite{trimmer2024artificial}. Here, we demonstrate ZodiAq's ability to navigate through an artificially created environment. The envisaged scenario is depicted in Figure \ref{fig::SoftInteraction}A, where the robot is commanded to follow a simple translation along the x-axis. Along the trajectory, an inclined obstruction is placed. The experiment aims to observe the strategy employed by this robot to intentionally interact and utilize the obstacle for its navigation. This approach contrasts the traditional strategy of mapping the environment and using advanced control algorithms to circumvent obstacles \cite{Cheng2021}.

\par
Figure \ref{fig::SoftInteraction}B shows the result of the experimental demonstration. The ZodiAq meets the rigid barrier at the 37-second mark. The robot's arm flexibility enables a safe and advantageous interaction, guiding movement along the barrier. Post-barrier, the ZodiAq resumes its intended path. Notably, at 70 seconds, the flagella presses against a cylindrical block, and at 78 seconds, against the wall's far edge, utilizing these interactions for forward thrust.

%
\subsection{Crawling Gait}
\begin{figure}
\centering
\includegraphics[width=1\columnwidth]{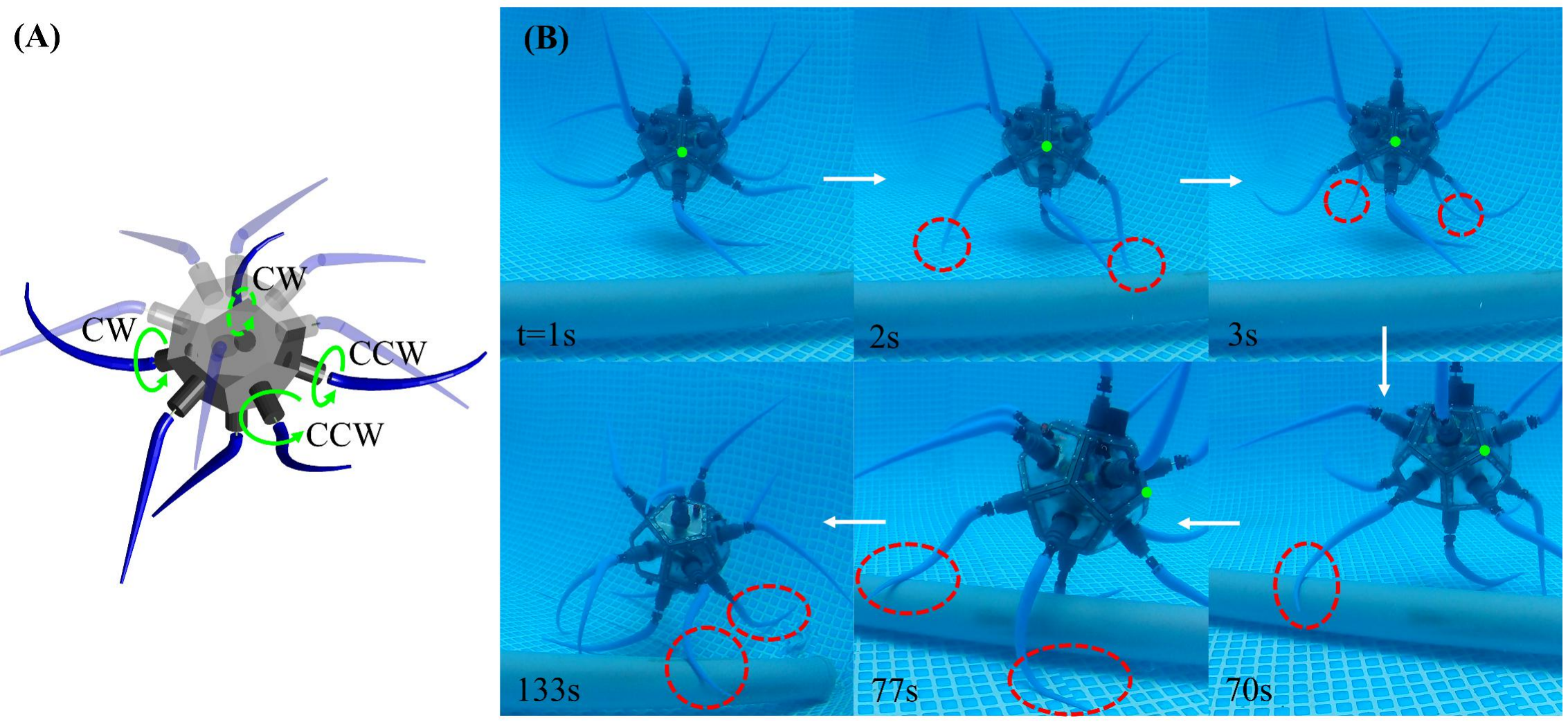}
\caption{Demonstration of ZodiAq’s crawling gait: (A) To achieve a turtle-like sweeping action, four actuators at the bottom are specifically actuated while keeping all other actuators passive. (B) Crawling locomotion of ZodiAq. Instances of contact with the ground, walls, and a cylindrical obstacle are evident in the video snapshots. Contact regions are demarcated with dashed ellipses. The green dot on a corner assists in visualizing the orientation of ZodiAq.
}
\label{fig::Crawling}
\end{figure}

In previous demonstrations, the locomotion of ZodiAq is achieved through the thrust generated by the flagella interacting with the surrounding fluid. Another form of propulsion is possible for this prototype: crawling. Crawling using appendages refers to a mode of locomotion where a robot moves using its limbs to push the body forward \cite{Arienti2013Oceans, Wu2021, Cianchetti2015}. Even though the prototype is not designed for crawling, we made it heavy enough to sink to the floor by increasing the ballast mass. Inspired by the movement of sea turtles, we implemented a manual open-loop actuation strategy (Figure \ref{fig::Crawling}A). To propel forward, the left bottom arms (M6 and M8) were rotated CCW, while the right side arms (M10 and M12) were rotated CW. All other actuators were kept passive.
\par
Figure \ref{fig::Crawling}B shows the crawling locomotion of ZodiAq. The arms perform a sweeping action that pushes the body forward. Additionally, the robot utilizes the environment, specifically the cylindrical obstacle, to aid its movement. We note that the locomotion cannot be classified as pure crawling, as it is also augmented by thrust from fluid interactions.
\subsection{Marine Exploration}
\begin{figure}
\centering
\includegraphics[width=1\columnwidth]{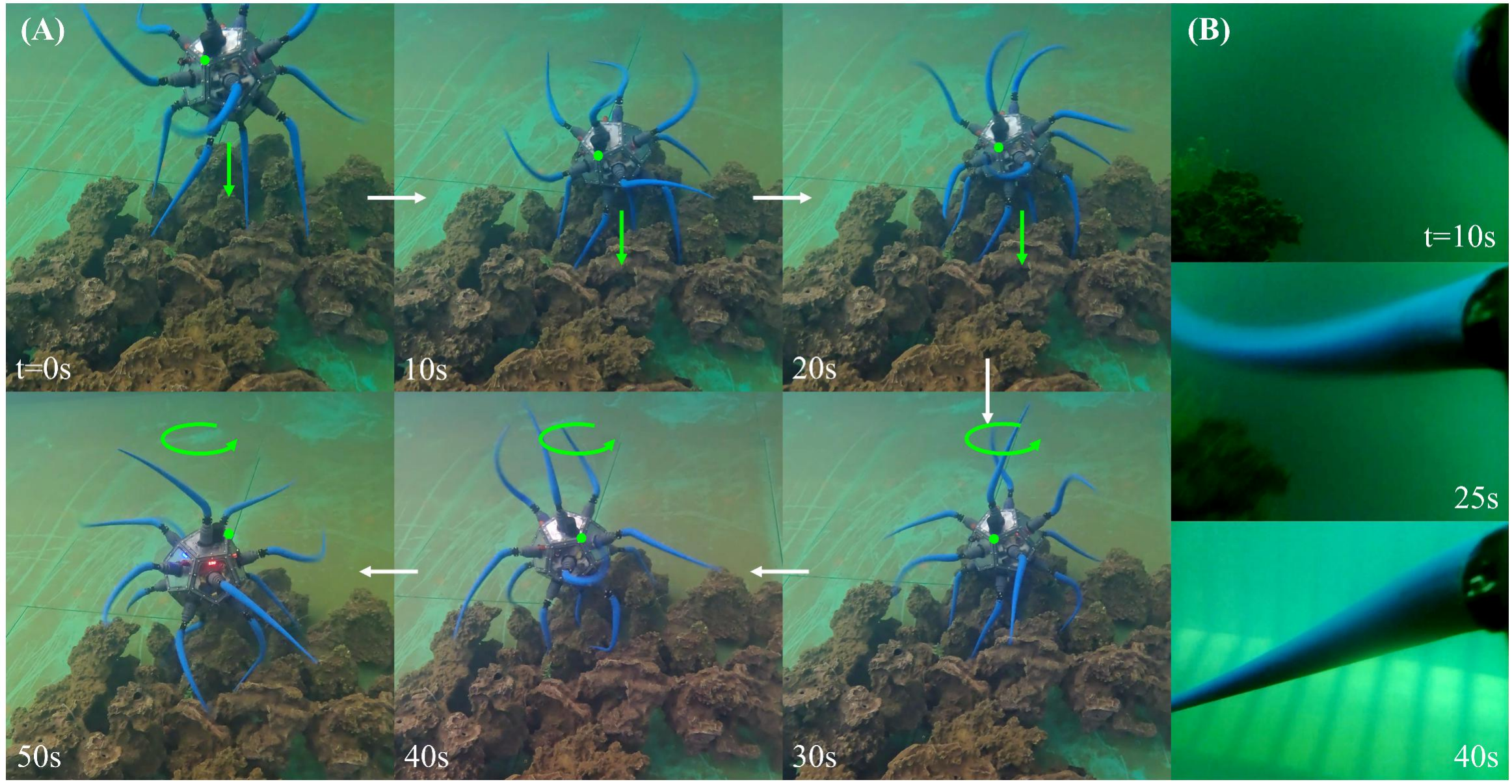}
\caption{Coral reef exploration with ZodiAq: (A) Snapshots of ZodiAq descending towards the coral reef and executing a rotation. The green dot on a corner assists in visualizing the orientation of ZodiAq. (B) Snapshots from the video footage captured by ZodiAq's onboard camera. 
}
\label{fig::Coral}
\end{figure}

Our final demonstration focuses on the application of the prototype for close inspection of a marine environment like that of a coral reef (Figure \ref{fig::Coral}). This scenario establishes a rocky underwater environment simulating a coral reef. As ZodiAq hovers above the target location, a descent command is sent using the acoustic modem. Upon receiving the command, the robot descends and contacts the coral. Despite this contact, it attempts to continue its downward motion. The compliance of the flagella ensures safe interaction with minimal control requirements. Subsequently, a command is sent to rotate the prototype to scan the environment. It can be seen that during the execution of the rotational movement, contact with the environment aids in facilitating the motion. 
\par
Figure \ref{fig::Coral}B shows the snapshots taken from the video footage recorded by the onboard camera. In the current prototype, a low-resolution PiCamera module is internally mounted behind a transparent face on one of the top faces. The quality of the footage is also affected by underwater low lighting conditions. This can be enhanced by integrating higher-quality external cameras or floodlights onto the prototype.

\section{Discussions and Conclusions}
\label{sec::conclusion}

In summary, the paper introduces the design of a novel soft underwater drone featuring a flagella-inspired swimming gait. We have developed a digital twin and a simplified control law to analyze and manage the motion of this complex system. The prototype has demonstrated motion capabilities in 4D: translation along all three axes and rotation about its vertical axis (effectively zero turning radius), which is advantageous for close inspection tasks. It has shown resilience in operating despite actuator impairments and has engaged in safe interactions with its surroundings, which are beneficial traits in unpredictable marine environments. The prototype demonstrates the ability to effectively utilize the environment to facilitate its motion with minimal control requirements, showcasing the embodied intelligence of its design.
%

Although the prototype features a range of innovative capabilities, there is scope for improvement in some aspects. The operational redundancy of the robot is constrained by the proposed control law. The control law predefines the rotation direction of the motors and combines the actuators on opposite faces to generate a new input that prevents the simultaneous movement of motors on opposite faces. This constraint is clearly visible in Figure \ref{fig::HeightOrientation}D, Figure \ref{fig::Redundancy}A, and Figure \ref{fig::Redundancy}B. A fault-tolerant control could be developed for robust navigation to prevent accidents after a motor failure occurs, taking advantage of the system's redundancy property. This opens up the possibility of developing advanced controllers to maintain control and stability in the system, even when failures occur in several defective motors.

While following a square trajectory, it was observed that design imperfections caused deviations in the -y direction movement, as illustrated in Figure \ref{fig::SquareTrajectory}A. These design imperfections can be incorporated into the control law by fine-tuning the thrust and moment vectors of individual actuators. These adjustments enable finer control, allowing the robot to accurately follow intended trajectories despite fabrication errors. Another obvious source of the deviation is the absence of feedback from planar coordinates in the experimental setup. Two or more acoustic modems can be implemented to triangulate the robot and provide planar coordinates feedback.

With these improvements, this prototype holds promise for real-world applications, signaling a significant step forward in the field of soft robotics for underwater exploration. The potential for these technologies to contribute to safe, efficient, and environmentally conscious exploration is immense, and we believe that our work lays a significant step toward realizing this potential. 

\section*{Acknowledgments}
The work was supported by the US Office of Naval Research Global under Grant N62909-21-1-2033 and in part by the Khalifa University of Science and Technology under Grants CIRA-2020-074 and RC1-2018-KUCARS-8474000136.



\listoffigures

\clearpage
\processdelayedfloats

\includepdf[pages=-]{SupplementaryMaterial.PDF}
\end{document}